\pdfoutput=1

\documentclass[11pt]{article}

\usepackage{acl}

\usepackage{times}
\usepackage{latexsym}

\usepackage[T1]{fontenc}

\usepackage[utf8]{inputenc}

\usepackage{microtype}

\usepackage{booktabs}
\usepackage{tabularx}
\usepackage{graphicx}
\usepackage{colortbl}
\usepackage{amssymb}
\usepackage{amsmath}

\usepackage{algorithm}
\usepackage{algpseudocode}
\usepackage{multirow}

\title{Cross-Platform and Cross-Domain Abusive Language Detection with Supervised Contrastive Learning}

\author{Md Tawkat Islam Khondaker$^{\dagger}$~~ { Muhammad Abdul-Mageed$^{\dagger}$}~~ { Laks V.S. Lakshmanan}
\\\\ 
\normalsize $^{\dagger}$Deep Learning \& Natural Language Processing Group  \\
\normalsize The University of British Columbia\\
\\
\texttt{\{tawkat@cs.,muhammad.mageed@,laks@cs.\}ubc.ca}
}

\begin{document}
\maketitle
\begin{abstract}
The prevalence of abusive language on different online platforms has been a major concern that raises the need for automated cross-platform abusive language detection. However, prior works focus on concatenating data from multiple platforms, inherently adopting Empirical Risk Minimization (ERM) method. In this work, we address this challenge from the perspective of domain generalization objective. We design SCL-Fish, a supervised contrastive learning integrated meta-learning algorithm to detect abusive language on unseen platforms. Our experimental analysis shows that SCL-Fish achieves better performance over ERM and the existing state-of-the-art models. We also show that SCL-Fish is data-efficient and achieves comparable performance with the large-scale pre-trained models upon finetuning for the abusive language detection task.
\end{abstract}

\section{Introduction}

Abusive langugage is defined as any form of microaggression, condescension, harassment, hate speech, trolling, and the like~\cite{jurgens}. Use of abusive language online has been a significant problem over the years. Although a plethora of works has explored automated detection of abusive language, it is still a challenging task due to its evolving nature~\cite{davidson,muller_2017,williams}. In addition, a standing challenging in tackling abusive language is linguistic variation as to how the problem manifests itself across different platforms~\cite{karan,swamy,salminen_2020}.

\begin{figure}[h]
  \includegraphics[width=\columnwidth]{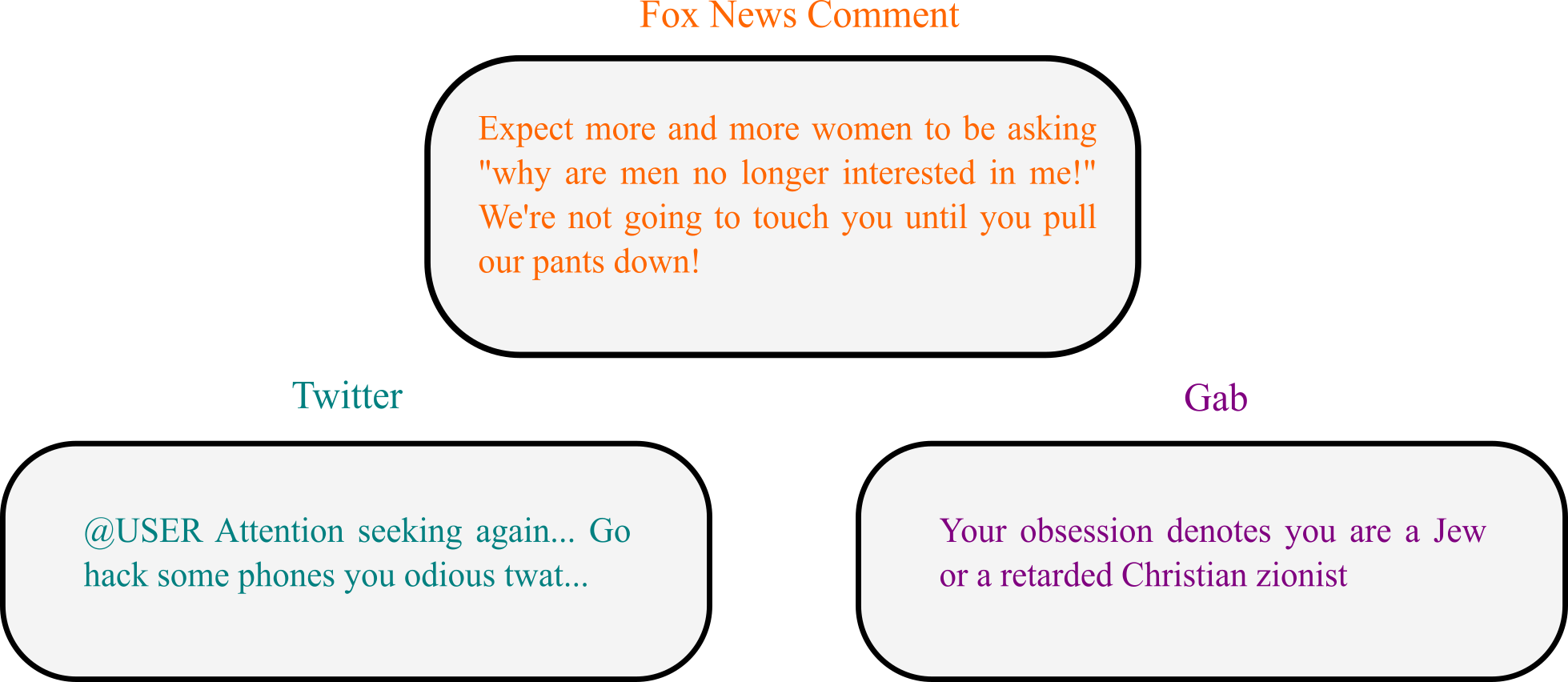}
  \caption{Examples of abusive language on different platforms.}
  \label{fig:examples}
\end{figure}

We provide examples illustrating variation of abusive language on different platforms in Figure~\ref{fig:examples}.\footnote{This paper contains several examples of abusive language and strong words for the purpose of demonstration.} For example, user comments in broadcasting media such as Fox News do not directly contain any strong words but can implicitly carry abusive messages. Meanwhile, people on social media such as on Twitter employ an abundance of strong words that can be outright personal bullying and spread of hate speech. On an extremist public forum such as Gab, users mostly spread abusive language in the form of identity attacks. For these reasons, it is an unrealistic assumption to train an abusive language detector on data from one platform and expect the model to exhibit equally satisfactory performance on another platform.

Prior Works on cross-platform abusive language detection~\cite{karan,mishra,corazza,salminen_2020} usually concatenate examples from multiple sources, thus inherently applying Empirical Risk Minimization (ERM)~\cite{erm}. These models capture platform-specific spurious features, and lack generalization~\cite{fish}.~\citet{fortuna}, on the other hand, incorporate out-of-platform data into training set and employ domain-adaptive techniques. Other works such as~\citet{swamy} and~\citet{gallacher} develop one model for each platform and ensemble them to improve overall performance. 

None of the prior works, however, attempt to generalize task-oriented features across the platforms to improve performance on an unseen platform. In this work, we introduce a novel method for learning domain-invariant features to fill this gap. 
Our approach initially adopts an first-order derivative of meta-learning algorithm~\cite{andrychowicz_maml,finn_maml}, \textit{Fish}~\cite{fish}, that attempts to capture domain-invariance. We then propose a supervised contrastive learning (\textit{SCL})~\cite{scl} to impose an additional constraint on capturing task-oriented features that can help the model to learn semantically effective embeddings by pulling samples from the same class close together while pushing samples from opposite classes further apart. We refer to our new method as \textbf{SCL-Fish} and conduct extensive experiments on a wide array of platforms representing social networks, public forums, broadcasting media, conversational chatbots, and synthetically-generated data to show the efficacy of our method over other abusive language detection models (and specially ERM that prior works on cross-platform abusive language detection applied).

To summarize, we offer the following contributions in this work:

\begin{enumerate}
    \item We propose SCL-Fish, a novel supervised contrastive learning augmented domain generalization method for cross-platform abusive language detection.
    \item Our method outperforms prior works on cross-platform abusive language detection, thus demonstrating superiority to ERM (the core idea behind these previous models). Additionally, we show that SCL-Fish outperforms platform-specific state-of-the-art abusive/hate speech detection models.
    \item Our analysis reveals that SCL-Fish can be data-efficient and exhibit comparable performance with the state-of-the-art models upon finetuning on the abusive language detection task.
\end{enumerate}

\section{Related Works}
\subsection{What is Abusive Language?}
The boundary between hate speech, offensive, and abusive language can be unclear.~\citet{davidson} define \textit{hate speech} as ``language that is used to express hatred towards a targeted group or is intended to be derogatory, to humiliate, or to insult the members of the group"; whereas,. ~\citet{olid} define \textit{offensive language} as ``any form of non-acceptable language (profanity) or a targeted offense, which can be veiled or direct". In this paper, we adopt the definition of abusive language provided by~\citet{jurgens} and consider both offensive and hate speech as abusive language in general, since distinguishing between offensive and hate speech is often deemed as subjective~\cite{sap,wilds}.

\subsection{Domain Generalization}

In the domain generalization task, training and test sets are sampled from different distributions~\cite{quinonero2008dataset}. 
In recent years, domain-shifted datasets have been introduced by synthetically corrupting the samples (\citealt{hendrycks2019robustness},~\citealt{xiao_madry},~\citealt{Santurkar}). To improve the capability of a learner on distributional generalization,~\citet{erm} proposes Empirical Risk Minimization (ERM) approach which is widely used as the standard for the domain generalization tasks (\citealt{wilds}). ERM concatenates data from all the domains and focuses on minimizing the average loss on the training set. However,~\citet{pezeshki} state that a learner can overestimate its performance by capturing only one or a few dominant features with the ERM approach.
Several other algorithms have been proposed to generalize models on unseen domains.~\citet{sagawa} attempt to develop distributionally robust algorithm, where the domain-wise losses are weighted inversely proportional to the domain performance.~\citet{krueger2021out} propose to minimize the variation loss across the domains during the training phase and~\citet{arjovskyinvariant} aim to penalize the models if the performance varies among the samples from the same domain.

\subsection{Contrastive Learning}

Contrastive learning aims to learn effective embedding by pulling semantically close neighbors together while pushing apart non-neighbors (\citealt{hadsell}). This method uses cross-entropy-based similarity objective to learn the embedding representation in the hyperspace~\cite{chen_2017,henderson_2017}. In computer vision,~\citet{chen_2020} proposes a framework for contrastive learning of visual representations without specialized architectures or a memory bank.~\citet{scl} shows that supervised contrastive loss can outperform cross-entropy loss on ImageNet~\cite{imagenet}. In NLP, similar methods have been explored in in the context of sentence representation learning~\cite{karpukhin,gillick,logeswaran}. Among of the most notable works,~\citet{simcse} proposes unsupervised contrastive learning framework, \textit{SimCSE} that predicts input sentence itself by augmenting it with dropout as noise.

\subsection{Abusive Language Detection}

Over the years, the task of abusive language detection have been studied in NLP in the form of hate speech~\cite{davidson,founta,golbeck}, sexism/racism~\cite{waseem}, cyberbulling~\cite{xu,dadvar}.
Earlier works in abusive language detection depend on feature-based approaches to identify lexical difference between abusive and non-abusive language~\cite{warner,waseem,ribeiro}. Although inclusion of neural network architecture improves the performance~\cite{mitrovic,kshirsagar,sigurbergsson}, the models still misclassify a large number of samples in false-positive and false-negative categories when abusive language is intentionally manipulated~\cite{gitari}. Recently, Transformer-based~\cite{transformer} architectures like BERT~\cite{bert}, RoBERTa~\cite{roberta} have been introduced in the abusive language detection task~\cite{nuli,swamy}.

However, most of the prior works on abusive language detection focus on a single platform due to the inaccessibility to multiple platforms~\cite{hatespeech_datasets} and thus, do not scale well on other platforms~\citet{schmidt}. As a result, the models are not suitable to apply to other platforms due to the lack of generalization~\cite{karan,grondahl}. In this work, we aim to address this challenge by introducing an augmented domain generalization method that captures task-oriented domain-generalized features across multiple platforms.

\section{Method}

\subsection{Challenge \& Proposed Solution}
\label{sec:challenge}

As shown in Figure~\ref{fig:examples}, the nature of offensive language can vary from one platform to another. Therefore, it is important to design a model that can capture platform-generalized representations. This inspires us to adopt a domain-generalization algorithm that can maximize feature generalization while avoiding dependence on domain-specific, spurious features. To learn platform-invariant features, we adopt first-order derivative of \textit{Inter-domain Gradient Matching (IDGM)}~\citet{fish}, a Model Agnostic Meta-Learning (MAML)~\cite{andrychowicz_maml,finn_maml}, algorithm, \textit{Fish}, that aims to reduce sample complexity of new, unseen domains and increase domain-generalized feature selection across those domains.

\begin{figure}[h]

  \includegraphics[width=\columnwidth]{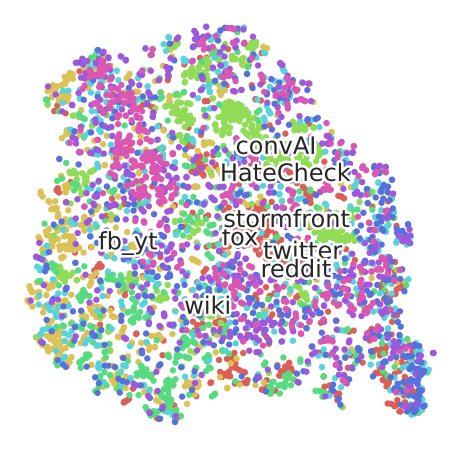}
  \caption{tSNE representations of platforms. We plot the embedding of [\texttt{CLS}] token from pre-trained BERT.}
  \label{fig:tsne_500}
\end{figure}

However, if we look at Figure~\ref{fig:tsne_500}, unlike the domain-generalization task, representation of abusive language across the platforms is more overlapping and scattered. Thus, the model should also learn some platform-specific and overlapping features that can help to capture task-oriented representations. Therefore, we need to impose a constraint on the learning objective of the model so that in one direction, it should learn platform-invariant features for better generalization, and in the other direction, it should also learn only those task-oriented overlapping features that pass positive signals to those platform-generalized features for the abusive language detection task.

To learn task-oriented features we introduce \textbf{SCL-Fish}, method for supervised contrastive learning (SCL)~\cite{scl} with Fish. The rationale behind integrating SCL is that we seek to find commonalities between the examples of each class (abusive/normal) irrespective of the platforms and contrast them with examples from the other class.

\subsection{SCL-Fish}

Assuming we have a training dataset of abusive language detection consisting of samples from two platforms $\mathbf{P_1}$ and $\mathbf{P_2}$ where $\mathbf{P_k}$ = ${\{(\mathbf{X}^k, \mathbf{Y}^k)\}}$. Given a model $\theta$ and loss function $\mathit{l}$, the empirical risk minimization (ERM)~\cite{erm} objective is to minimize the average loss across the given platform:
\begin{equation*}
\begin{aligned}[b]
    L_{ERM} = \min_{\theta}\;\mathbb{E}_{\;(x,y)\;\sim\;P\in({P_1},\;{P_2})} \frac{\delta\mathit{l}((x,\;y);\;\theta)}{\delta\theta}
\end{aligned}
\end{equation*}

The expected gradients for these two platforms are expressed as
\begin{gather*}
    G_1 \;=\; \mathbb{E}_{(x,y)\;\sim\;{P_1}} \frac{\delta\mathit{l}\,((x,\;y)\;;\;\theta)}{\delta\theta}\\
    G_2 \;=\; \mathbb{E}_{(x,y)\;\sim\;{P_2}} \frac{\delta\mathit{l}\,((x,\;y)\;;\;\theta)}{\delta\theta}
\end{gather*}

If the directions of $G_1$ and $G_2$ are same ($G_1$.$G_2$ > 0), then we can say that the model is improving on both platforms. Therefore, IDGM algorithm attempts to align the direction of the gradients $G_1$ and $G_2$ by maximizing their inner dot product. Hence, given the total number of training platforms $S$, the final objective function of IDGM is obtained by subtracting gradient dot product (GIP) from ERM loss:
\begin{equation}
\begin{aligned}[b]
    & L_{IDGM} = L_{ERM} \;\\
    & - \; \gamma\,\frac{2}{S(S\,-\,1)} \sum_{i,j \in S}^{i \neq j}G_i.G_j
\end{aligned}
\end{equation}

Here, $\gamma$ is a scaling term and GIP can be computed in linear time by $\widehat{G}$ = $||\sum_i\,G_i||^2$ - $\sum_i||\,G_i||^2$

However, the derivation of $\widehat{G}$ is computationally expensive, as it is a dot product of two gradients. Adopting from \citet{reptile}, \citet{fish} work around this issue by proposing a first-order derivative version of IDGM, namely, Fish. \citet{fish} show that given the gradient of ERM $\Bar{G}$ and a clone of original model $\Tilde{\theta}$,
\\

$G_f$ = $\mathbb{E}[\theta\:-\:\Tilde{\theta}]$ - $\alpha\,S.\Bar{G}$ and $G_g$ = $\frac{d\,\widehat{G}}{d\,\theta}$,
\begin{equation}
    \lim_{\alpha \rightarrow 0} \; \frac{G_f.G_g}{||G_f||\;.\;||G_g||} \; = \; 1
\end{equation}

In other words, if we ignore the ERM objective, we can substitute the second-order derivative $G_g$ with a computationally less expensive $G_f$.

Although, this method exhibits impressive performance on the domain-generalization task, as mentioned in Section~\ref{sec:challenge}, it may capture only platform-invariant features without much focus on \textit{task-relevant} features. To overcome this issue, we augment Fish with a supervised contrastive learning (SCL) objective, which will teach the model to select the features such that the representation of an abusive sample and a non-abusive sample are located far from each other in the hyperspace,
\begin{equation}
\begin{aligned}[b]
\label{eq:scl}
    & L_{SCL} = - \sum_{j=1}^N\;1_{y_i=y_j}\\
    & \log\frac{exp(f(x_i)\,.\,f(x_j)\,/\,\tau)}{\sum1_{i\neq k}\:exp(f(x_i)\,.\,f(x_k)\,/\,\tau)}
\end{aligned}
\end{equation}

Here, \textit{f}(.) is an encoder and \textit{N} is the number of samples summing all the platforms. Therefore, the model will be encouraged to learn only those task-oriented features that are invariant across the platforms \textit{and} can be used to distinguish abusive and non-abusive examples.

\begin{algorithm}[h]
\caption{SCL-Fish}\label{algo:scl_fish}

\begin{algorithmic}[1]

\For {iteration = 1, 2,...}
\State $\Tilde{\theta} \gets \theta$
    \For {$P_i \in \{P_1, P_2, ..., P_S\}$} 
        \State Sample minibatch $p_i \sim P_i$
        \State $\Tilde{g_i} = \mathbb{E}_{(x,y) \sim p_i} \left[ \frac{\delta\mathit{l}\,((x,\;y)\;;\;\Tilde{\theta})}{\delta\Tilde{\theta}} \right]$
        \\
        \State Update $\Tilde{\theta} \gets \Tilde{\theta} - \alpha\Tilde{g_i}$ 
    \EndFor
    \\
    \State Update $\theta \gets \theta -  \epsilon(\Tilde{\theta} - \theta)$   \Comment{Updating Fish}
    \\
    \State $P_{scl} \gets \{P_1 \cup P_2 \cup ... \cup P_S\}$
    \For {Sample minibatch $p_{scl} \sim P_{scl}$}
    
        \State \Comment{Calculate gradient for SCL from \eqref{eq:scl}:}
        
        \State $g_{scl} = \mathbb{E}_{(x,y) \sim p_{scl}} \left[ \frac{\delta\mathit{l}\,((x,\;y)\;;\;\Tilde{\theta})}{\delta\Tilde{\theta}} \right]$
    
        \\
        \State Update $\theta \gets \theta - \alpha'\:g_{scl}$ 
    \EndFor

\EndFor

\end{algorithmic}
\end{algorithm}

\begin{table}[t]
\centering
\footnotesize  
\scalebox{0.8}{
\begin{tabular}{lllr}
\hline
\textbf{Dataset} & \textbf{Platform}                                              & \textbf{Source}                                                                                                                                                                                                                                                                                                                                                                                                                 & \textbf{Offnsv/normal} \\ \hline
wiki             & Wikipedia                                                      & \citet{wulczyn}                                                                                                                                                                                                                                                                                                                                                                                            & 14880 / 117935                      \\

twitter          & Twitter                                                        & \textit{Multiple*} & 77656 / 55159  
\\

fb-yt           & \begin{tabular}[c]{@{}c@{}}Facebook \\\& Youtube\end{tabular}  & \citet{salminen}                                                                                                                                                                                                                                                                                                                                                                                           & 2364 / 858                          \\
stormfront       & Stormfront                                                     & \citet{gibert}                                                                                                                                                                                                                                                                                                                                                                                            & 1364 / 9507                         \\
fox              & Fox News                                                      & \citet{gao}                                                                                                                                                                                                                                                                                                                                                                                                 & 435 / 1093                          \\
twi-fb          & \begin{tabular}[c]{@{}c@{}}Twitter \&\\ Facebook\end{tabular}  & \citet{mandl}                                                                                                                                                                                                                                                                                                                                                                                              & 6840 / 11491                        \\
reddit           & Reddit                                                         & \citet{qian}                                                                                                                                                                                                                                                                                                                                                                                               & 2511 / 11073                        \\
convAI           & \begin{tabular}[c]{@{}c@{}}ELIZA \&\\ CarbonBot\end{tabular}   & \citet{curry}                                                                                                                                                                                                                                                                                                                                                                                             & 128 / 725                           \\
hateCheck        & \begin{tabular}[c]{@{}c@{}}Synthetic.\\ Generated\end{tabular} & \citet{rottger}                                                                                                                                                                                                                                                                                                                                                                                          & 2563 / 1165                         \\
gab              & Gab                                                            & \citet{qian}                                                                                                                                                                                                                                                                                                                                                                                               & 15270 / 656                         \\
yt\_reddit       & \begin{tabular}[c]{@{}c@{}}Youtube \\\& Reddit\end{tabular}    & \citet{mollas}                                                                                                                                                                                                                                                                                                                                                                                           & 163 / 163                           \\ \hline
\end{tabular}
}
\caption{\label{table:datasets}
List of experimental datasets with corresponding platforms. \textbf{*} \emph{Twitter} dataset is collected from~\citet{waseem},~\citet{davidson},~\citet{jha},~\citet{elsherief},~\citet{founta},~\citet{mathur},~\citet{basile},~\citet{mandl},~\citet{ousidhoum}, and~\citet{olid}.
}
\end{table}

We present SCL-Fish in Algorithm~\ref{algo:scl_fish}. For each training platform, Fish performs inner-loop (\textit{l3}-\textit{l8}) update steps with learning rate $\alpha$ on a clone of the original model $\Tilde{\theta}$ in a minibatch. Subsequently, the original model $\theta$ is updated by a weighted difference between the cloned model and the original model $\Tilde{\theta} \: - \: \theta$. After performing, platform-generalized update, the trained samples of this iteration(\textit{l12}) are queued and sampled in a minibatch to update $\theta$ with supervised contrastive loss (\textit{l13}-\textit{l18}).

\section{Experiments}

\subsection{Datasets}

To experiment with the efficacy of SCL-Fish, we compile datasets from a wide range of platforms. We collect source of the datasets primarily from~\cite{unified_datasets} and~\cite{hatespeech_datasets}. We provide meta-information of the datasets in Table~\ref{table:datasets}. Description of each dataset is presented in Appendix~\ref{appendix:datasets}.

\subsection{Methods Comparison}

We compare performance of \textbf{SCL-Fish} with \textbf{Fish}, also using \textbf{ERM} as a sensible baseline. We also conduct experiments on an SCL version of ERM \textbf{(SCL-ERM)}. Additionally, we compare SCL-Fish with two of the benchmark models for abusive/hate speech detection, HateXplain~\cite{hatexplain} and HateBERT~\cite{hatebert}. \textbf{HateXplain} is finetuned on hate speech detection datasets collected from Twitter and Gab\footnote{https://gab.com} for a three-class classification (hate, offensive, or normal) task. It incorporates human-annotated explainability with BERT to gain better performance by reducing unintended bias towards target communities. While conducting our experiments, we consider both \textit{hate} and \textit{offensive} classes as one category (\textit{abusive}). \textbf{HateBERT} pre-trains BERT with Masked Language Modeling (MLM) objective on more than one million offensive and hate messages from banned Reddit community. It results in a shifted BERT model that has learned language variety and hate polarity (e.g. \textit{hate}, \textit{abuse}). Finetuning on different abusive language detection tasks has shown that HateBERT achieves the best/comparable performance.

\begin{table*}[t]
\centering
\resizebox{\linewidth}{!}{
\begin{tabular}{@{}clccccccccccccccccccccccc@{}}
\toprule
\textbf{Platform}                                                    &  & \multicolumn{3}{c}{\textbf{HateXplain}}                                                                                                  & \multicolumn{1}{l}{} & \multicolumn{3}{c}{\textbf{HateBERT}}                                                                                                    & \multicolumn{1}{l}{} & \multicolumn{3}{c}{\textbf{ERM}}                                                                                                         & \multicolumn{1}{l}{} & \multicolumn{3}{c}{\textbf{SCL-ERM}}                                                                                                     & \multicolumn{1}{l}{} & \multicolumn{3}{c}{\textbf{Fish}}                                                                                                       & \multicolumn{1}{l}{} & \multicolumn{3}{c}{\textbf{SCL-Fish}}                                                                                                    \\ \cmidrule(r){1-1} \cmidrule(lr){3-5} \cmidrule(lr){7-9} \cmidrule(lr){11-13} \cmidrule(lr){15-17} \cmidrule(lr){19-21} \cmidrule(l){23-25} 
\textbf{(\% of hate)}                                                &  & \textbf{Acc}  & \textbf{\begin{tabular}[c]{@{}c@{}}Pos.\\ F\textsubscript{1}\end{tabular}} & \textbf{\begin{tabular}[c]{@{}c@{}}Macro\\ F\textsubscript{1}\end{tabular}} & \multicolumn{1}{l}{} & \textbf{Acc}  & \textbf{\begin{tabular}[c]{@{}c@{}}Pos.\\ F\textsubscript{1}\end{tabular}} & \textbf{\begin{tabular}[c]{@{}c@{}}Macro\\ F\textsubscript{1}\end{tabular}} & \multicolumn{1}{l}{} & \textbf{Acc}  & \textbf{\begin{tabular}[c]{@{}c@{}}Pos\\ F\textsubscript{1}\end{tabular}} & \textbf{\begin{tabular}[c]{@{}c@{}}Macro\\ F\textsubscript{1}\end{tabular}} & \multicolumn{1}{l}{} & \textbf{Acc}  & \textbf{\begin{tabular}[c]{@{}c@{}}Pos.\\ F\textsubscript{1}\end{tabular}} & \textbf{\begin{tabular}[c]{@{}c@{}}Macro\\ F\textsubscript{1}\end{tabular}} & \multicolumn{1}{l}{} & \textbf{Acc} & \textbf{\begin{tabular}[c]{@{}c@{}}Pos.\\ F\textsubscript{1}\end{tabular}} & \textbf{\begin{tabular}[c]{@{}c@{}}Macro\\ F\textsubscript{1}\end{tabular}} & \multicolumn{1}{l}{} & \textbf{Acc}  & \textbf{\begin{tabular}[c]{@{}c@{}}Pos.\\ F\textsubscript{1}\end{tabular}} & \textbf{\begin{tabular}[c]{@{}c@{}}Macro\\ F\textsubscript{1}\end{tabular}} \\ \midrule
\begin{tabular}[c]{@{}c@{}}\textbf{stormfront}\\ (12.5)\end{tabular} &  & \textbf{88.1} & 44.1                                                       & 67.2                                                        &                      & 87.3          & 34.6                                                       & 63.8                                                        &                      & 85.3          & \textbf{44.2}                                              & 67.7                                                        & \textbf{}            & 86.0          & 43.0                                                       & 67.5                                                        &                      & 85.5         & 42.0                                                       & 66.9                                                        &                      & 85.1          & \textbf{44.2}                                              & \textbf{67.8}                                               \\
\begin{tabular}[c]{@{}c@{}}\textbf{fox}\\ (28.5)\end{tabular}        &  & \textbf{73.9} & 29.4                                                       & 56.7                                                        &                      & 68.7          & 31.5                                                       & 63.8                                                        &                      & 73.6          & 42.3                                                       & 62.6                                                        & \textbf{}            & 73.6          & 42.3                                                       & 62.6                                                        &                      & 73.6         & 44.3                                                       & 63.5                                                        &                      & 72.2          & \textbf{47.5}                                              & \textbf{64.3}                                               \\
\begin{tabular}[c]{@{}c@{}}\textbf{twi-fb}\\ (37.3)\end{tabular}        &  & \cellcolor[HTML]{E8E8E8}63.4 & \cellcolor[HTML]{E8E8E8}09.3                                                       & \cellcolor[HTML]{E8E8E8}43.2                                                        &                      & \cellcolor[HTML]{E8E8E8}\textbf{65.0}          & \cellcolor[HTML]{E8E8E8}27.9                                                       & \cellcolor[HTML]{E8E8E8}52.4                                                        &                      & \cellcolor[HTML]{E8E8E8}61.3          & \cellcolor[HTML]{E8E8E8}35.7                                                       & \cellcolor[HTML]{E8E8E8}54.0                                                      &            & \cellcolor[HTML]{E8E8E8}60.2          & \cellcolor[HTML]{E8E8E8}33.6                                                       & \cellcolor[HTML]{E8E8E8}52.6                                                    &                      & \cellcolor[HTML]{E8E8E8}53.7         & \cellcolor[HTML]{E8E8E8}36.9                                                       & \cellcolor[HTML]{E8E8E8}50.2                                                     &                      & \cellcolor[HTML]{E8E8E8}61.8          & \cellcolor[HTML]{E8E8E8}\textbf{38.2}                                              & \cellcolor[HTML]{E8E8E8}\textbf{55.3}                                               \\
\begin{tabular}[c]{@{}c@{}}\textbf{reddit}\\ (18.5)\end{tabular}     &  & \textbf{83.7} & 38.0                                                       & 64.3                                                        &                      & \cellcolor[HTML]{E8E8E8}81.0          & \cellcolor[HTML]{E8E8E8}45.5                                                       & \cellcolor[HTML]{E8E8E8}\textbf{66.9}                                               &                      & 76.9          & 43.0                                                       & 64.3                                                        & \textbf{}            & 77.7          & 43.9                                                       & 65.1                                                        &                      & 76.7        & 44.6                                                        & 64.9                                                        &                      & 76.6          & \textbf{46.3}                                              & 65.7                                                        \\
\begin{tabular}[c]{@{}c@{}}\textbf{convAI}\\ (15.0)\end{tabular}     &  & 86.4          & 26.6                                                       & 59.5                                                        &                      & \textbf{87.9} & 56.9                                                       & 74.9                                                        &                      & 86.6          & 66.3                                                       & 78.9                                                        & \textbf{}            & 86.8          & 65.9                                                       & 78.8                                                        &                      & 86.3         & 64.7                                                       & 78.1                                                        &                      & 87.3          & \textbf{67.7}                                              & \textbf{79.9}                                               \\
\begin{tabular}[c]{@{}c@{}}\textbf{hateCheck}\\ (68.8)\end{tabular}  &  & 38.4          & 26.9                                                       & 36.9                                                        &                      & 58.9          & 64.3                                                       & 57.9                                                        &                      & \textbf{67.3} & \textbf{77.4}                                              & 59.0                                                        & \textbf{}            & 65.4          & 75.3                                                       & 58.6                                                        &                      & 67.1         & 76.6                                                       & \textbf{60.5}                                                        &                      & 66.7          & 76.2                                                       & 60.4                                               \\
\begin{tabular}[c]{@{}c@{}}\textbf{gab}\\ (95.9)\end{tabular}        &  & \cellcolor[HTML]{E8E8E8}75.6          & \cellcolor[HTML]{E8E8E8}85.7                                                      & \cellcolor[HTML]{E8E8E8}50.6                                               &                      & 75.9          & 86.0                                                       & 50.4                                                        &                      & 91.1          & 95.3                                                       & \textbf{59.1}                                                        & \textbf{}            & 91.4          & 95.5                                                       & 57.9                                                        &                      & 90.9        & 95.2                                                        & 58.8                                                       &                      & \textbf{92.0} & \textbf{95.8}                                              & 57.4                                                        \\
\begin{tabular}[c]{@{}c@{}}\textbf{yt-reddit}\\ (50.0)\end{tabular}  &  & 65.3          & 54.3                                                       & 63.2                                                        &                      & 70.9          & 69.3                                                       & 70.8                                                        &                      & 72.4          & 75.7                                                       & 71.9                                                        & \textbf{}            & \textbf{74.5} & \textbf{77.1}                                              & \textbf{74.2}                                               &                      & 73.6         & 76.6                                                      & 73.2                                                         &                      & 73.0          & 76.7                                                       & 72.3                                                        \\ \midrule
\textbf{avg.}                                                        &  & 71.9          & 38.9                                                       & 55.2                                                        &                      & 74.5          & 52.0                                                       & 61.6                                                        &                      & 76.8         & 59.9                                                        & 64.7                                                        & \textbf{}            & \textbf{76.9} & 59.6                                                       & 64.7                                                        &                      & 75.9         & 60.1                                                      & 64.5                                                         &                      & 76.8* & \textbf{61.6}                                              & \textbf{65.4}                                               \\ \bottomrule
\end{tabular}
}
\caption{
\label{table:cross_platform}
Performance on cross-platform datasets. \textbf{Bold} font represents the best performance for a particular metric. \textcolor[HTML]{808080}{Gray} cells indicate performance on the datasets from identical or overlapping platforms but different sources and distributions. \textbf{*} Although SCL-Fish exhibits comparable accuracy with other competitive models on this imbalanced dataset, it achieves better accuracy on the balanced dataset (Appendix~\ref{appendix:cross_platform_balanced}).
}
\end{table*}

\subsection{Experimental Setup} 

We train the models (ERM, SCL-ERM, Fish, and SCL-Fish) on \emph{fb-yt}, \emph{twitter}, and \emph{wiki} datasets (in-platform datasets) and use \emph{stromfront} as validation set. We use the same hyperparameters on all the models for fair comparisons. We present the list of hyperparameters in Appendix~\ref{appendix:hyperparameters}. The rest of the datasets from Table~\ref{table:datasets} are used for cross-platform evaluation. As evident from Table~\ref{table:datasets}, the datasets are highly imbalanced. Hence, we report \textit{F\textsubscript{1}-score} for abusive class (we denote it as \textit{positive-F\textsubscript{1}}) and \textit{macro-averaged F\textsubscript{1}-score}. For completeness, we also provide performance in \textit{accuracy}. We train and evaluate our models on Nvidia A100 40GB GPU.

\section{Results on Cross-Platform Datasets}

We show results of our models for cross-platform performance in Table~\ref{table:cross_platform}. We observe that SCL-Fish outperforms other methods in macro-F\textsubscript{1} and positive-F\textsubscript{1} scores while maintaining comparable performance with the best method on the other datasets (\emph{reddit}, \emph{hatecheck}).
In overall average performance, SCL-Fish achieves best macro-F\textsubscript{1} and positive-F\textsubscript{1} scores. More specifically, user comments on broadcasting media (\textit{Fox News}), SCL-Fish achieves a gain of $3.2$\% positive-F\textsubscript{1} and $0.5$\% macro-F\textsubscript{1} over the other methods. On public forums (\textit{Youtube} and \textit{Reddit}), SCL-Fish achieves a total gain of $2.0$\% in positive-F\textsubscript{1} but SCL-ERM outperforms SCL-Fish by $1.3$\% in macro-F\textsubscript{1} score. On AI bot conversation (\textit{CarbonBot} and \textit{ELIZA}), SCL-Fish achieves a gain of $1.4$\% positive-F\textsubscript{1} and $1.0$\% macro-F\textsubscript{1} over other methods. On the synthetically-generated platform (\textit{HateCheck}), ERM outperforms SCL-Fish by $1.2$\% in positive-F\textsubscript{1} score and Fish outperforms SCL-Fish by $0.1$\% in macro-F\textsubscript{1} score. On Gab, all the methods (ERM and Fish-based, including SCL-Fish) achieve high positive-F\textsubscript{1} score because of the highly imbalanced dataset. Hence, for a fair comparison among all methods, we report performance on sampled balanced datasets in Appendix~\ref{appendix:cross_platform_balanced}. We also discuss the performance on the in-platform datasets in Appendix~\ref{appendix:in_platform}.

Most notably, HateBERT achieves the highest macro-F\textsubscript{1} scores on \emph{reddit}, which is expected since HateBERT is pre-trained on \emph{reddit} and so has an advantage over other methods since these are trained on data from other platforms. However, all the models including HateXplain and HateBERT are trained on the datasets from Twitter platform. Hence, we analyze performance of the models on \emph{twi-fb} dataset. Our rationale is that although twi-fb involves data from Twitter and Facebook, these data do not necessarily have the same distribution as data used to train all the models. The distribution of datasets from the same platform can still defer due to the variations in the timestamps, topics, locations, demographic attributes (e.g. age, race, gender, ethnicity). Although it is not possible to extract all this information from the textual contents, we provide a quantitative comparison between in-domain and out-domain datasets for Twitter in Appendix~\ref{appendix:distribution_comparison}. We refer the readers to \citet{wilds} for more detailed analysis. We find that performance of the models deteriorates significantly (under $56$\% macro-F\textsubscript{1}) even on datasets from overlapping platforms but of different distributions. This demonstrates effect of distribution shift in the data, even if we train on date from the same platform. 
We further discuss possible rationales for this performance gap across the platforms in Appendix~\ref{appendix:word_freq}.

\section{Analysis}

In this section, we conduct qualitative and quantitative analysis on the experimental results.

\subsection{Diversity over Quantity}

It is worth noting that HateBERT has been pre-trained on $1,478,348$ Reddit messages, almost five times more data than SCL-Fish. However, as Table~\ref{table:cross_platform} shows, performance of HateBERT on cross-platform datasets suffers significant drops which is not the case for SCL-Fish. Even on \emph{yt-reddit} dataset, which is collected from \textit{Youtube} and \textit{Reddit} (the latter being the platform whose data HateBERT is trained on), HateBERT fails to outperform the baseline ERM method. This shows that, for the purpose of  creating platform/domain-invariant models, it is more important to employ training data with different distributions than simply using huge amounts of training data from the same platform but that may have limited distribution.


\subsection{Finetuning SCL-Fish}
\label{sec:finetuning}

Since we show SCL-Fish exhibits better performance than other methods on most of the cross-platform datasets, we further investigate whether the platform-generalization capability of SCL-Fish helps it improve performance on a specific platform (\textit{Twitter}) upon finetuning. For this purpose, we use two benchmark datasets, namely, OLID~\cite{olid} dataset from SemEval-2019 Task 6~\cite{semeval_19} and AbusEval~\cite{abuseval}. Please note that we use OLID dataset for training our methods (Appendix~\ref{appendix:datasets}). Now we are finetuning with the same dataset for this experiment.

\begin{table}[h]

\resizebox{\columnwidth}{!}{
\begin{tabular}{ccccccc}
\hline
\textbf{Datasets}                    &  & \textbf{Models}                   &           & \textbf{\begin{tabular}[c]{@{}c@{}}Macro\\ F\textsubscript{1}\end{tabular}} & \textbf{} & \textbf{\begin{tabular}[c]{@{}c@{}}Pos.\\ F\textsubscript{1}\end{tabular}} \\ \hline
\multirow{4}{*}{\textbf{OffensEval}} &  & BERT                              &           & 80.3                                                        &           & 71.5                                                       \\
                                     &  & HateBERT                          &           & 80.9                                                        &           & 72.3                                                       \\
                                     &  & NULI                              &           & \textbf{82.9}                                               &           & \textbf{75.2}                                              \\
                                     &  & SCL-Fish                          &           & \underline{81.6}                                                        &           & \underline{72.6}                                                       \\ \hline
\multirow{4}{*}{\textbf{AbusEval}}   &  & BERT                              &           & 72.7                                                        &           & 55.2                                                       \\
                                     &  & HateBERT                          &           & \textbf{76.5}                                               &           & \textbf{62.3}                                              \\
                                     &  & \citet{abuseval} & \textbf{} & 71.6                                                        &           & 53.1                                                       \\
                                     &  & SCL-Fish                          &           & \underline{75.2}                                                        &           & \underline{59.4}                                                       \\ \hline 
\end{tabular}
}
\caption{ \label{table:finetuning}
Performance of models after finetuning. \textbf{Bold} and \underline{\textit{underline}} represent best and second best performance for a particular metric, respectively.
}
\end{table}

We present results for this set of experiments in Table~\ref{table:finetuning}. Performance of NULI (BERT-based model secured first position in SemEval-2019 Task 6~\cite{semeval_19}) in the table is from~\citet{nuli} and BERT, HateBERT from~\citet{hatebert}.

As Table~\ref{table:finetuning} shows, NULI~\cite{nuli} achieves the best performance for OLID dataset. Although SCL-Fish gets a lower score than NULI\footnote{Please note that~\citet{hatebert} reports positive-F\textsubscript{1} of NULI as $59.9$\% which is lower than positive-F\textsubscript{1} of SCL-Fish. But the positive-F\textsubscript{1} we compute from~\citet{nuli} is different from the one reported in~\citet{hatebert}. Therefore, we consider our computed positive-F\textsubscript{1} for NULI.}, SCL-Fish outperforms BERT and HateBERT on both in positive-F\textsubscript{1} and macro-F\textsubscript{1}. This is important because HateBERT uses five times more data from one specific platform (\textit{Reddit}). This proves that our proposed SCL-Fish is useful not only in platform generalized zero-shot setting but also for finetuning, and emphasizes the importance of \textit{diversity} of the data (which translates into varied distributions) over data \textit{size}.

For AbusEval dataset, SCL-Fish performs better than BERT and the prior work~\cite{abuseval}, but it cannot outperforms HateBERT. We suspect that the reason is due to the different annotation process followed during the earlier training phase of SCL-Fish and HateBERT. Because, although OLID and AbusEval contain identical tweets in the training and the testing sets, the annotation scheme of AbuseEval is different from OLID. While~\citet{olid} uses the definition of offensive language as ``Posts containing any form of non-acceptable language (profanity) or a targeted offense, which can be veiled or direct" to annotate OLID dataset,~\citet{abuseval} uses the definition of abusive language as ``hurtful language that a speaker uses to insult or offend another individual or a group of individuals based on their personal qualities, appearance, social status, opinions, statements, or actions" to annotate AbusEval dataset. More comprehensively, AbusEval excludes any kind of untargeted messages from the hate speech category. During the training phase of SCL-Fish, we consider any targeted or non-targeted strong language as offensive. Therefore, finetuning on AbusEval causes misalignment with the earlier training phase of SCL-Fish, and may result in performance deterioration.

\subsection{Explainability with Attention Visualization}

\begin{figure}[h]
  \includegraphics[width=\columnwidth]{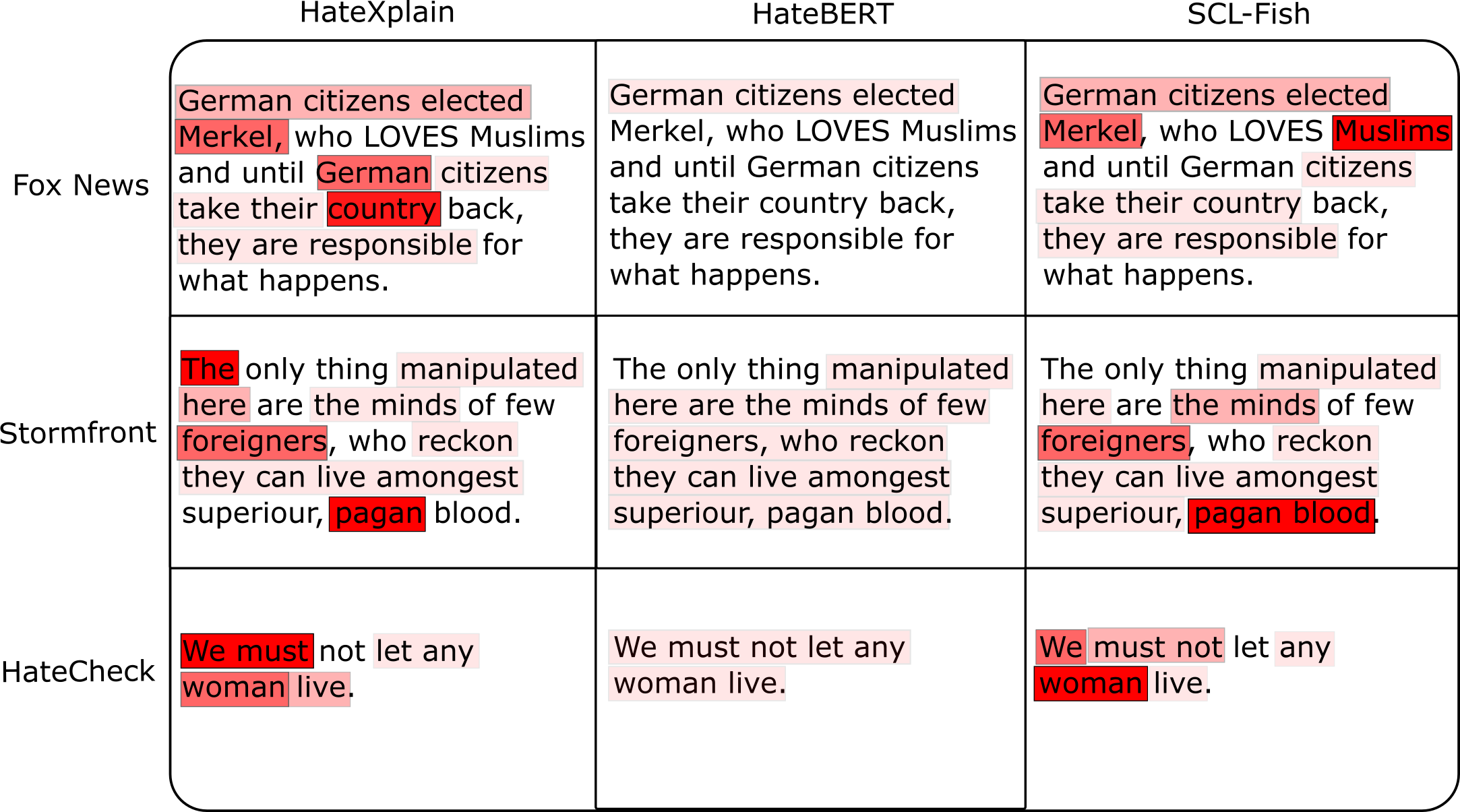}
  \caption{Attention visualization for different platforms. Deeper color indicates higher attention.}
  \label{fig:attention}
\end{figure}

We investigate how platform generalization helps the model attend to the right context on `out-of-platform' datasets. For this purpose, we analyze attention vectors of SCL-Fish, HateXplain, and HateBERT in an attempt to better understand their performance. We use BertViz~\cite{bertviz} to compute and visualize the final layer attention vectors from \texttt{[CLS]} to other tokens. We select three out-of-platform datasets (\emph{fox}, \emph{stormfront}, and \emph{hateCheck}) and randomly sample one abusive example from each where SCL-Fish correctly identifies the example as abusive, but HateXplain and HateBERT misclassify it. Figure~\ref{fig:attention} shows the attention visualization for each of the examples. As we can see, in the example from Fox News user comments, although the text does not explicitly contain any strong or offensive words, it is seemingly offensive towards `Muslims' and `Merkel'. Hence, our models should attend to these two words with the highest priority, which SCL-Fish does. On the other hand, although HateXplain gives higher attention to `Merkel', it fails to attend the word `Muslims'. Surprisingly, HateBERT does not assign priority to any context for the misclassified examples. On the example from StormFront, both SCL-Fish and HateXplain, correctly assign priority to the words `foreigners' and `pegan' unlike HateBERT. However, HateXplain also confuses other words e.g. `The' as a highly prioritized token. Finally, the example from synthetically-generated dataset \emph{hateCheck} is challenging because of the linguistic complexity (e.g. negations, hedging terms) language models typically struggle to address~\cite{hossain_negation,bert_not,kassner_negated}. We observe that SCL-Fish highly prioritizes `women' and also attends to the token `not'. On the other hand, HateXplain mistakenly provides the highest attention to `We must' and ignores the negation term `not'.

Overall, our analysis shows that model trained on platform-generalized settings improves on identifying the targeted community and right context on an out-domain offensive text. On the contrary, platform-specific models may not be able to attend to the targeted community in a different platform, because these models are trained on target specific to particular platforms.

\subsection{SCL Improves Fish}

\begin{figure}[h]
  \includegraphics[width=\columnwidth]{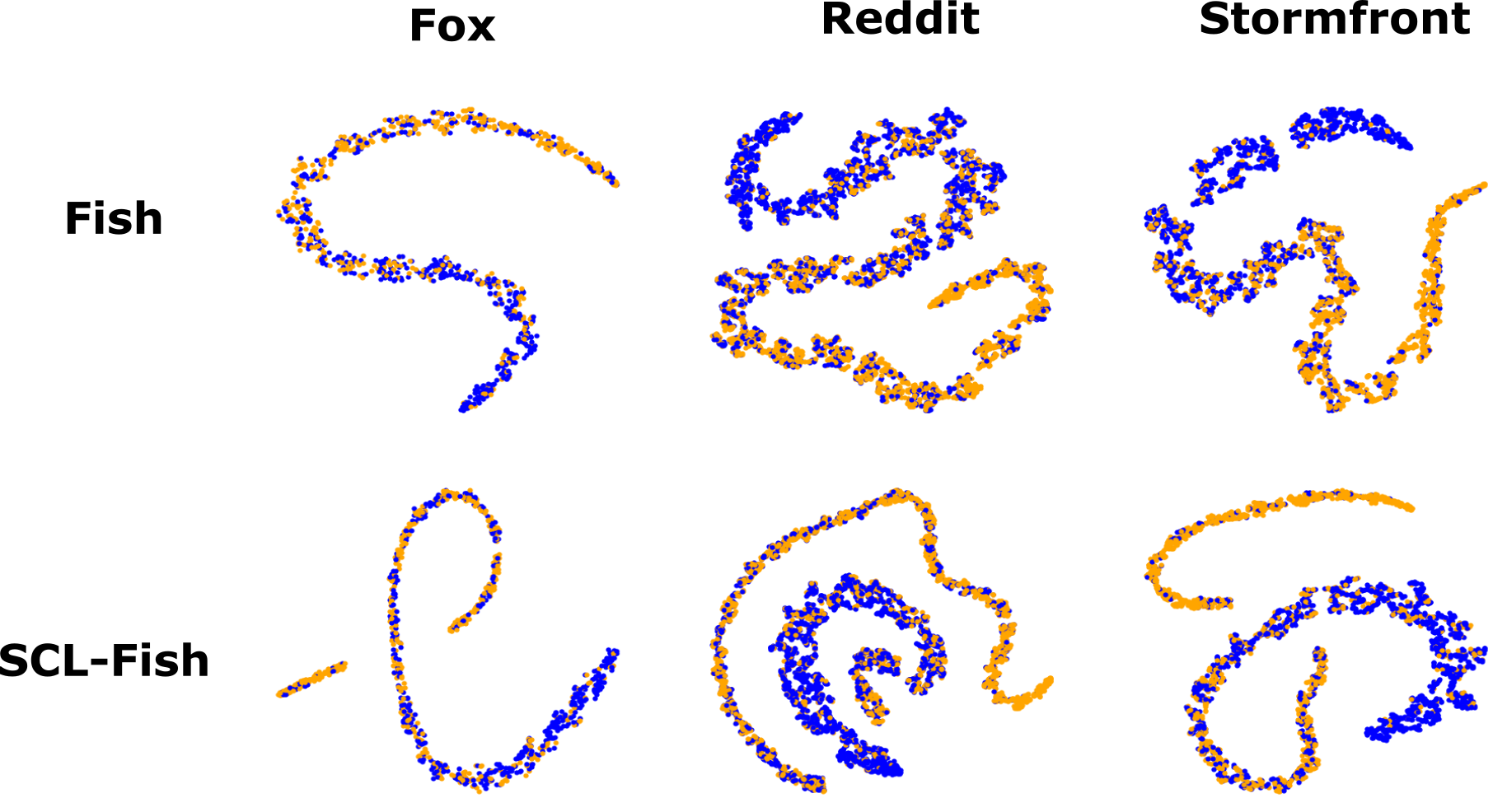}
  \caption{tSNE plot for Fish vs. SCL-Fish on Fox News Comment, Reddit, and StormFront. Abusive samples are presented in \textcolor{orange}{\textit{orange}} and non-abusive samples are presented in \textcolor{blue}{\textit{blue}}.}
  \label{fig:tsne_fish_scl_fish}
\end{figure}

From Table~\ref{table:cross_platform} and Table~\ref{table:in_platform}, it is evident that integrating SCL with Fish empirically improves performance across the platforms. Now, we substantiate the empirical result with the visual justification for Fish and SCL-Fish on different platforms. For all the platforms, we pass an equal number of abusive and non-abusive samples to the models and plot the [\texttt{CLS}] embeddings in Figure~\ref{fig:tsne_fish_scl_fish}.

We observe that, SCL-Fish forms more compact clusters of abusive (majority from \textcolor{orange}{\textit{orange}} samples) and non-abusive (majority from \textcolor{blue}{\textit{blue}} samples) examples than Fish. Supervised contrastive learning attempts to learn task-oriented features that help bring representations of the same class closer to each other while pushing representations of different classes further apart. As a result, distinct clusters are formed for each class in Figure~\ref{fig:tsne_fish_scl_fish}. Therefore, incorporating SCL helps Fish reduce the confusion between abusive and non-abusive representations and improves overall performance of the model.

\section{Limitations}

Although SCL-Fish achieves improvement over Fish, training SCL-Fish takes longer time than Fish. Empirically, we find that SCL-Fish is approximately 1.2x slower than Fish. Moreover, we believe that the subjective nature of abusive language~\cite{sap} affects the annotation process of different datasets and possibly negatively impact performance. We conduct an error analysis in Appendix~\ref{appendix:error_analysis}.

\section{Conclusion}

In this work, we addressed the problem of cross-platform abusive language detection from the domain generalization perspective. We proposed SCL-Fish, a supervised contrastive learning augmented meta-learning method to learn generalized task-driven features across platforms. We showed that SCL-Fish achieves better performance compared to the other state-of-the-art models and models adopting ERM for cross-platform abusive language detection. Our analysis also reveals that SCL-Fish achieves comparable performance on finetuning with much smaller data for cross-platform training than other data-intensive methods. Our work demonstrates progress on both platform and domain generalization in the context of abusive language detection, which we hope future research can be extended to other areas of language understanding.


\bibliography{anthology,custom}
\bibliographystyle{acl_natbib}

\appendix

\section{Hyperparameter Configuration}
\label{appendix:hyperparameters}

The detailed configuration of hyperparameters for the training phase of the cross-platform experiments is shown in Table~\ref{tab:hyperp}. We run each experiment three times and report the average performance of the models.

\begin{table}[H]
\centering
\begin{tabular}{@{}ll@{}}
\toprule
\textbf{Hyperparameters} & \textbf{Values} \\ \midrule
LM model variant         & BERT-base-uncased    \\
Token length             & 512           \\
Optimizer                & Adam           \\
AdamW epsilon            & 1e-8            \\
AdamW betas              & (0.9, 0.999)    \\
Fish meta lr. ($\epsilon$) & 0.05            \\
SCL temperature ($\tau$)   & 0.05            \\
Learning rate            & 5e-6            \\
Batch size               & 8              \\
Epochs                   & 10              \\ \bottomrule
\end{tabular}
\caption{Hyperparameters for cross-platform experiments.}
\label{tab:hyperp}
\end{table}

Table~\ref{tab:hyperp_finetuning} presents the configuration of hyperparameters during the finetuning (Section~\ref{sec:finetuning}).

\begin{table}[H]
\centering
\begin{tabular}{@{}ll@{}}
\toprule
\textbf{Hyperparameters} & \textbf{Values} \\ \midrule
LM model variant         & BERT-base-uncased    \\
Token length             & 100            \\
Optimizer                & AdamW           \\
AdamW epsilon            & 1e-8            \\
AdamW betas              & (0.9, 0.999)    \\
Learning rate            & 1e-5            \\
Batch size               & 32              \\
Epochs                   & 5              \\ \bottomrule
\end{tabular}
\caption{Hyperparameters for finetuning.}
\label{tab:hyperp_finetuning}
\end{table}

\section{Performance on Cross-Platform Balanced Datasets}
\label{appendix:cross_platform_balanced}

\begin{table*}[h]
\centering
\resizebox{\linewidth}{!}{
\begin{tabular}{cccccccccccccccccccccccccc}
\hline
                                    & \textbf{}            & \multicolumn{3}{c}{\textbf{HateXplain}}                                                                                                                 & \textbf{} & \multicolumn{3}{c}{\textbf{HateBERT}}                                                                                                                   & \textbf{} & \textbf{} & \multicolumn{3}{c}{\textbf{ERM}}                                                                                                                        & \textbf{} & \multicolumn{3}{c}{\textbf{SCL-ERM}}                                                                                                                    & \textbf{} & \multicolumn{3}{c}{\textbf{Fish}}                                                                                                                       & \textbf{} & \multicolumn{3}{c}{\textbf{SCL-Fish}}                                                                                                                            \\ \cline{3-5} \cline{7-9} \cline{12-14} \cline{16-18} \cline{20-22} \cline{24-26} 
\multirow{-2}{*}{\textbf{Platform}} & \textbf{}            & \textbf{Acc}                 & \textbf{\begin{tabular}[c]{@{}c@{}}Pos.\\ F\textsubscript{1}\end{tabular}} & \textbf{\begin{tabular}[c]{@{}c@{}}Macro\\ F\textsubscript{1}\end{tabular}} & \textbf{} & \textbf{Acc}                 & \textbf{\begin{tabular}[c]{@{}c@{}}Pos.\\ F\textsubscript{1}\end{tabular}} & \textbf{\begin{tabular}[c]{@{}c@{}}Macro\\ F\textsubscript{1}\end{tabular}} & \textbf{} & \textbf{} & \textbf{Acc}                 & \textbf{\begin{tabular}[c]{@{}c@{}}Pos.\\ F\textsubscript{1}\end{tabular}} & \textbf{\begin{tabular}[c]{@{}c@{}}Macro\\ F\textsubscript{1}\end{tabular}} & \textbf{} & \textbf{Acc}                 & \textbf{\begin{tabular}[c]{@{}c@{}}Pos.\\ F\textsubscript{1}\end{tabular}} & \textbf{\begin{tabular}[c]{@{}c@{}}Macro\\ F\textsubscript{1}\end{tabular}} & \textbf{} & \textbf{Acc}                 & \textbf{\begin{tabular}[c]{@{}c@{}}Pos.\\ F\textsubscript{1}\end{tabular}} & \textbf{\begin{tabular}[c]{@{}c@{}}Macro\\ F\textsubscript{1}\end{tabular}} & \textbf{} & \textbf{Acc}                          & \textbf{\begin{tabular}[c]{@{}c@{}}Pos.\\ F\textsubscript{1}\end{tabular}} & \textbf{\begin{tabular}[c]{@{}c@{}}Macro\\ F\textsubscript{1}\end{tabular}} \\ \hline
\textbf{stormfront}                 &                      & 64.7                         & 48.5                                                       & 60.9                                                        &           & 61.9                         & 41.2                                                       & 56.5                                                        &           &           & 69.2                         & 60.1                                                       & 67.5                                                        &           & 67.5                         & 56.4                                                       & 65.3                                                        &           & 67.3                         & 56.1                                                       & 65.0                                                        &           & \textbf{69.5}                         & \textbf{60.6}                                              & \textbf{67.8}                                               \\
\textbf{fox}                        &                      & 57.0                         & 30.7                                                       & 49.8                                                        &           & 55.6                         & 36.3                                                       & 51.1                                                        &           &           & 61.5                         & 46.9                                                       & 58.4                                                        &           & 61.6                         & 46.9                                                       & 58.4                                                        &           & 61.8                         & 49.1                                                       & 59.3                                                        &           & \textbf{63.3}                         & \textbf{54.6}                                              & \textbf{61.9}                                               \\
\textbf{twi-fb}                    &                      & \cellcolor[HTML]{E8E8E8}51.6 & \cellcolor[HTML]{E8E8E8}09.4                               & \cellcolor[HTML]{E8E8E8}38.2                                &           & \cellcolor[HTML]{E8E8E8}55.5 & \cellcolor[HTML]{E8E8E8}29.0                               & \cellcolor[HTML]{E8E8E8}48.3                                &           &           & \cellcolor[HTML]{E8E8E8}54.7 & \cellcolor[HTML]{E8E8E8}38.9                               & \cellcolor[HTML]{E8E8E8}51.5                                &           & \cellcolor[HTML]{E8E8E8}53.4 & \cellcolor[HTML]{E8E8E8}36.6                               & \cellcolor[HTML]{E8E8E8}49.9                                &           & \cellcolor[HTML]{E8E8E8}50.0 & \cellcolor[HTML]{E8E8E8}\textbf{42.1}                      & \cellcolor[HTML]{E8E8E8}49.1                                &           & \cellcolor[HTML]{E8E8E8}\textbf{55.8} & \cellcolor[HTML]{E8E8E8}41.6                               & \cellcolor[HTML]{E8E8E8}\textbf{53.0}                       \\
\textbf{reddit}                     &                      & 61.6                         & 41.4                                                       & 56.4                                                        &           & \cellcolor[HTML]{E8E8E8}66.1 & \cellcolor[HTML]{E8E8E8}55.8                               & \cellcolor[HTML]{E8E8E8}64.2                                &           &           & 65.9                         & 57.9                                                       & 64.6                                                        &           & 66.1                         & 58.1                                                       & 64.8                                                        &           & 67.1                         & 60.6                                                       & 66.2                                                        &           & \textbf{68.2}                         & \textbf{63.2}                                              & \textbf{67.6}                                               \\
\textbf{convAI}                     &                      & 57.8                         & 28.0                                                       & 49.1                                                        &           & 73.4                         & 66.7                                                       & 72.3                                                        &           &           & 87.1                         & 87.2                                                       & 87.1                                                        &           & 86.3                         & 86.2                                                       & 86.3                                                        &           & 85.5                         & 85.3                                                       & 85.5                                                        &           & \textbf{87.9}                         & \textbf{87.9}                                              & \textbf{87.9}                                               \\
\textbf{hateCheck}                  &                      & 52.3                         & 27.5                                                       & 45.9                                                        &           & \textbf{63.4}                & 60.9                                                       & \textbf{63.3}                                               &           &           & 59.5                         & \textbf{67.3}                                              & 57.1                                                        &           & 59.1                         & 65.7                                                       & 57.6                                                        &           & 60.9                         & 67.1                                                       & 59.5                                                        &           & 60.8                                  & 66.8                                                       & 59.5                                                        \\
\textbf{gab}                        &                      & \cellcolor[HTML]{E8E8E8}33.8 & \cellcolor[HTML]{E8E8E8}41.0                               & \cellcolor[HTML]{E8E8E8}32.8                                &           & 33.9                         & 42.7                                                       & 32.3                                                        &           &           & 64.1                         & 72.8                                                       & \textbf{60.1}                                               &           & 62.2                         & 72.1                                                       & 56.7                                                        &           & \textbf{64.3}                & \textbf{72.9}                                              & \textbf{60.1}                                               &           & 60.2                                  & 71.1                                                       & 53.6                                                        \\
\textbf{yt-reddit}                  & \multicolumn{1}{l}{} & 65.3                         & 54.3                                                       & 63.2                                                        &           & 70.9                         & 69.3                                                       & 70.8                                                        &           &           & 72.4                         & 75.7                                                       & 71.9                                                        & \textbf{} & \textbf{74.5}                & \textbf{77.1}                                              & \textbf{74.2}                                               &           & 73.6                         & 76.6                                                       & 73.2                                                        &           & 73.0                                  & 76.7                                                       & 72.3                                                        \\ \hline
\textbf{avg.}                       &                      & 55.5                         & 35.1                                                       & 49.5                                                        &           & 60.1                         & 50.2                                                       & 57.4                                                        &           &           & 66.8                         & 63.3                                                       & 64.8                                                        &           & 66.4                         & 62.4                                                       & 64.2                                                        &           & 66.3                         & 63.7                                                       & 64.7                                                        &           & \textbf{67.3}                         & \textbf{65.3}                                              & \textbf{65.5}                                               \\ \hline
\end{tabular}
}
\caption{
\label{table:cross_platform_balanced}
Performance on the \textbf{balanced} cross-platform datasets. \textbf{Bold} font represents best performance for a particular metric. \textcolor[HTML]{808080}{Gray} cells indicates performance on the datasets from identical or overlapping platforms but different sources and distributions.
}
\end{table*}

We sample an equal number of examples from abusive and normal classes for each dataset. The result is shown in Table~\ref{table:cross_platform_balanced}.

\section{In-Platform Performance}
\label{appendix:in_platform}
\begin{table*}[h]
\centering
\resizebox{\linewidth}{!}{
\begin{tabular}{ccccccccccccccccc}
\hline
\textbf{Platform}                                                 &  & \multicolumn{3}{c}{\textbf{ERM}}                                                                                                         &           & \multicolumn{3}{c}{\textbf{SCL-ERM}}                                                                                                     &  & \multicolumn{3}{c}{\textbf{Fish}}                                                                                                       &  & \multicolumn{3}{c}{\textbf{SCL-Fish}}                                                                                                    \\ \cline{1-1} \cline{3-5} \cline{7-9} \cline{11-13} \cline{15-17} 
\textbf{(\% of hate)}                                             &  & \textbf{Acc}  & \textbf{\begin{tabular}[c]{@{}c@{}}Pos.\\ F\textsubscript{1}\end{tabular}} & \textbf{\begin{tabular}[c]{@{}c@{}}Macro\\ F\textsubscript{1}\end{tabular}} &           & \textbf{Acc}  & \textbf{\begin{tabular}[c]{@{}c@{}}Pos.\\ F\textsubscript{1}\end{tabular}} & \textbf{\begin{tabular}[c]{@{}c@{}}Macro\\ F\textsubscript{1}\end{tabular}} &  & \textbf{Acc} & \textbf{\begin{tabular}[c]{@{}c@{}}Pos.\\ F\textsubscript{1}\end{tabular}} & \textbf{\begin{tabular}[c]{@{}c@{}}Macro\\ F\textsubscript{1}\end{tabular}} &  & \textbf{Acc}  & \textbf{\begin{tabular}[c]{@{}c@{}}Pos.\\ F\textsubscript{1}\end{tabular}} & \textbf{\begin{tabular}[c]{@{}c@{}}Macro\\ F\textsubscript{1}\end{tabular}} \\ \hline
\textbf{\begin{tabular}[c]{@{}c@{}}fb-yt\\ (73.4)\end{tabular}}   &  & \textbf{94.1} & \textbf{95.8}                                              & \textbf{92.9}                                               & \textbf{} & 92.9          & 94.9                                                       & 91.4                                                        &  & 79.9         & 85.1                                                       & 77.1                                                        &  & 90.1          & 92.8                                                       & 88.5                                                        \\
\textbf{\begin{tabular}[c]{@{}c@{}}twitter\\ (58.5)\end{tabular}} &  & \textbf{89.2} & 90.7                                                       & \textbf{88.9}                                               & \textbf{} & \textbf{89.2} & \textbf{90.8}                                              & 88.8                                                        &  & 84.0         & 85.8                                                       & 83.8                                                        &  & \textbf{89.2} & 90.7                                                       & \textbf{88.9}                                               \\
\textbf{\begin{tabular}[c]{@{}c@{}}wiki\\ (11.2)\end{tabular}}    &  & \textbf{96.2} & \textbf{83.5}                                              & \textbf{90.7}                                               & \textbf{} & 96.0          & 82.2                                                       & 89.9                                                        &  & 95.1         & 77.9                                                       & 87.6                                                        &  & 95.9          & 82.7                                                       & 90.2                                                        \\ \hline
\textbf{avg.}                                                     &  & \textbf{93.2} & \textbf{90.0}                                              & \textbf{90.8}                                               & \textbf{} & 92.7          & 89.3                                                       & 90.1                                                        &  & 86.3         & 82.9                                                       & 82.8                                                        &  & 91.8          & 88.7                                                       & 89.2                                                        \\ \hline
\end{tabular}
}
\caption{\label{table:in_platform}
Performance on in-platform datasets. \textbf{Bold} font represents best performance for a particular metric.
}
\end{table*}

Table~\ref{table:in_platform} shows the performance of the methods on the in-platform datasets. Unsurprisingly, ERM-based methods outperform Fish-based methods on all the datasets and in all the metrics. ERM method learns platform-specific features, while the Fish-based method tends to learn platform-invariant features. Therefore, evaluating the in-platform datasets yield better performance for ERM-based methods. Notably, as the percentage of abusive speech decreases from the top row to the bottom row in Table~\ref{table:in_platform}, positive-F\textsubscript{1} scores also drop accordingly. But Fish-based methods suffer least performance deterioration ($10.1$\% drop from \emph{fb-yt} to \emph{wiki} for SCL-Fish, $7.2$\% drop from \emph{fb-yt} to \emph{wiki} for Fish) than the other methods ($12.3$\% drop from \emph{fb-yt} to \emph{wiki} for ERM, $12.7$\% drop from \emph{fb-yt} to \emph{wiki} for SCL-ERM). This shows that domain generalization helps the methods to learn more robust platform-invariant features, which in turn, results in more accurate detection of abusive speech on cross-platform datasets.

\section{Quantitative Comparison for \emph{Twitter} In-Domain and Out-Domain Datasets}
\label{appendix:distribution_comparison}

We compare \emph{twitter} (in-domain) and \emph{twi-fb} (out-domain) datasets based on linguistic features and sentiment analysis. For each dataset, we compute average sentiment scores, average number of words, and characters for both abusive and non-abusive classes.

\begin{table}[h]

\resizebox{\columnwidth}{!}{
\begin{tabular}{llrr}
\hline
Class                                                                            & \textbf{Features}                                           & \textbf{\emph{twitter}} & \textbf{\emph{twi-fb}} \\ \hline
\multirow{3}{*}{\textbf{Abusive}}                                                & \begin{tabular}[c]{@{}l@{}}No. of\\ Words\end{tabular}      & 15.49            & 29.64           \\
                                                                                 & \begin{tabular}[c]{@{}l@{}}No. of\\ Characters\end{tabular} & 96.53            & 168.46          \\
                                                                                 & \begin{tabular}[c]{@{}l@{}}Sentiment\\ Score\end{tabular}   & -0.75            & -0.83           \\ \hline
\multirow{3}{*}{\textbf{\begin{tabular}[c]{@{}l@{}}Non-\\ Abusive\end{tabular}}} & \begin{tabular}[c]{@{}l@{}}No. of\\ Words\end{tabular}      & 18.51            & 26.84           \\
                                                                                 & \begin{tabular}[c]{@{}l@{}}No. of\\ Characters\end{tabular} & 118.41           & 172.09          \\
                                                                                 & \begin{tabular}[c]{@{}l@{}}Sentiment\\ Score\end{tabular}   & -0.49            & -0.71           \\ \hline
\end{tabular}
}
\caption{\label{table:distribution_comparison}
Comparison between in-domain (\emph{twitter}) and out-domain (\emph{twi-fb}) datasets. Features are computed averaging the examples for a particular class (abusive/non-abusive).
}
\end{table}

Table~\ref{table:distribution_comparison} reflects the difference in sentiments scores and linguistic features between the datasets. We see that the number of words and the number of characters are higher for the out-domain (\emph{twi-fb}) dataset than the in-domain (\emph{twitter}) dataset for both abusive and non-abusive classes. Additionally, the examples of out-domain datasets have more negative sentiment on average than the examples of in-domain dataset. These types of variation can shift the distribution of the datasets, as a result, the models may struggle to perform better on an out-domain dataset (Table~\ref{table:cross_platform}).

\section{Rationale for Performance Gap across Platforms}
\label{appendix:word_freq}

\begin{figure*}[t]
  \includegraphics[width=\linewidth]{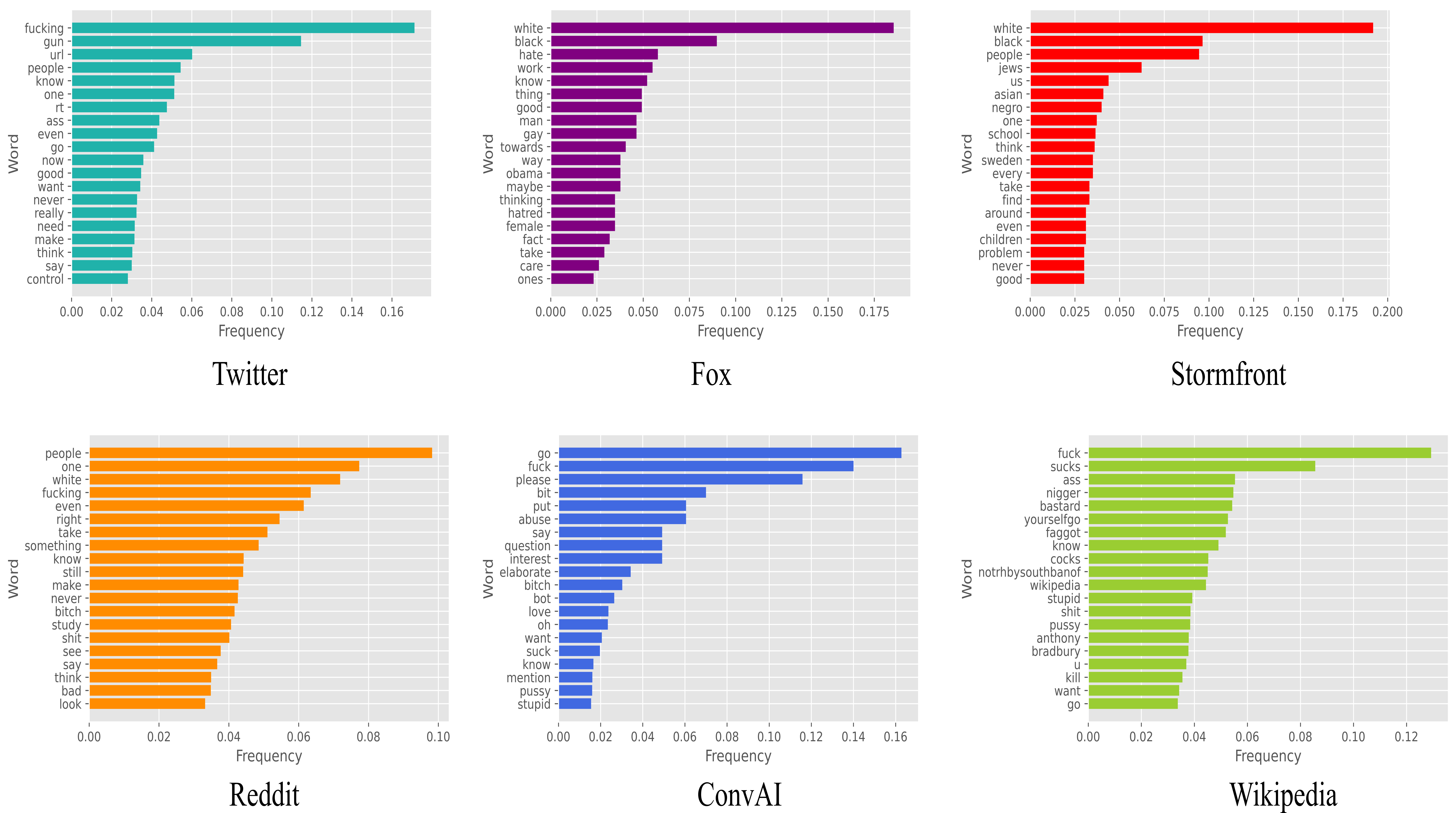}
  \caption{Top-20 normalized word frequency of abusive language for different platforms (ignoring \textit{stopwords} and \textit{non-alphabetic} characters).}
  \label{fig:word_freq}
\end{figure*}

To this end, we aim to study the reason for the performance gap of the models across different platforms through a qualitative analysis of linguistic variance. We sample abusive texts from the platforms and plot the word frequency in Figure~\ref{fig:word_freq}.

We observe that the type of abusive texts varies along with the linguistic features across the platforms. For example, on social networks like Twitter, most appeared words in abusive texts are `f*cking', `gun', `a*s', which mostly imply violence and personal attack. Meanwhile, an extremist forum like Stormfront contains words like `black', `white', `jews' which indicate abusive comments towards a particular community or ethnicity. Linguistic features from a public forum like Reddit reveal that abusive comments on this platform are mostly targeted attacks and slang. Abusive conversation with AI bots mostly contains strong words in the form of personal attacks. On the other hand, user comments on broadcasting media like Fox News do not contain any strong words but rather implicit abuse focused towards a particular race like `black', person like `Obama', or sexual orientations like `gay'. Finally, abusive texts on Wikipedia include both targeted and untargeted slang words toward a specific entity.

The variation of abuse across different platforms shows that training models on a specific platform are not enough to address the issue of mitigating abusive language on another platform. This also implies the importance of the platform-generalized study of abusive language detection.

\section{Datasets Description}
\label{appendix:datasets}

In this section, we briefly describe the datasets we compile for our cross-platform experiments.

\subsection{wiki}

\emph{wiki} dataset represents Wikipedia platform. We collect this dataset from~\citet{wulczyn}. The corpus contains 63M comments from discussions relating to user pages and articles dating from 2004 to 2015. Human annotations were used to label personal attack, aggressiveness, and harassment. The authors find that almost $1$\% of Wikipedia comments contain personal attacks. We randomly sample 132,815 examples from the initial corpus to make it compatible in size with other training sets. Of these examples, 14,880 contain abusive (personal attack, aggressiveness, harassment) language.

\subsection{twitter}

We collect \emph{twitter} dataset from a variety of sources.~\citet{waseem} annotate around 16k tweets that contain sexist/racist language. Initially, the authors bootstrap the corpus based on common slurs, then manually annotate the whole corpus to identify tweets that are offensive but do not contain any slur. Similarly,~\citet{davidson} crawled tweets with lexicon containing words and phrases identified by internet users as hate speech. Then crowdsourcing is performed to distinguish the category of hate, offensive, and normal tweets, resulting in around 25k annotated tweets.~\citet{jha} crawled Twitter with the terms that generally exhibit positive sentiment but sexist in nature (e.g. `as good
as a man', `like a man', `for a girl'). The authors also annotate tweets that are aggressively sexist. The final corpus contains around 10k tweets of implicit/explicit sexist and normal tweets.~\citet{elsherief} adopt multi-step data collection process that include collecting tweets based on lexicon, \textit{hashtag}, and other existing works~\cite{waseem,davidson}. Then, crowdsourcing is applied to annotate targeted and untargeted hate speech.~\citet{founta} build an annotated corpus of 80k tweets with seven classes (offensive, abusive, hateful speech, aggressive, cyberbullying, spam, and normal).~\citet{mathur} annotate a corpus of around 3k tweets containing hate, abusive, and normal tweets.~\citet{basile} crawled 13k tweets containing abusive language against women and immigrants. The authors applied crowdsourcing to annotate if the tweets contain individual/ group hate speech or aggressiveness.~\citet{mandl} develop a corpus of 7k English examples with the category of hate, offensive, and profanity.~\citet{ousidhoum} build a corpus of multilingual and multi-aspect hate speech. The English corpus (5,647 tweets) covers a wide range of hate speech categories including the level of directness, hostility, targeted theme, and targeted group.~\citet{olid} develop an offensive corpus of 14,100 tweets based on hierarchical modelings, such as whether a tweet is offensive/targeted, if it is targeted towards a group or individual.

Our final \emph{twitter} dataset contains 132.815 examples of which 77,656 are abusive.

\subsection{fb-yt}

\emph{fb-yt} represent both Facebook and Youtube platforms. We collect this dataset from~\citet{salminen}.~\citet{salminen} crawled the comments from Facebook and Youtube videos and annotate them into hateful, non-hateful categories. The authors also subcategorize hateful comments into 21 classes including accusation, promoting violence, and humiliation.

\subsection{stormfront}

\emph{stormfront} dataset is collected from~\citet{gibert}. The authors crawled around 10k examples from Stormfront and categorize them into hate/normal speech. The authors further investigate whether joining subsequent seemingly normal sentences result in hate speech. Our final dataset contains 1364 hateful speech from Stormfront.

\subsection{fox}

\emph{fox} dataset represents user comments on the broadcasting platform Fox News. We collect this dataset from~\citet{gao}. The authors find that the hateful comments are more implicit and creative and such hateful comments detection requires context-dependency.

\subsection{twi-fb}

\emph{twi-fb} dataset contains user posts from Twitter and Facebook. We collect this dataset from~\citet{mandl}. The authors initially collect the corpus by crawling keywords and hashtags. Later, they annotate the corpus into targeted/untargeted hate speech, offense, and profane.

\subsection{reddit}

\emph{reddit} dataset contains conversations from Reddit.~\citet{qian} compiled a list of toxic subreddit and crawled user conversations from those subreddits. Additionally, the authors provide hate speech intervention, where the goal is to automatically generate responses to intervene during online conversations that contain hate speech. The final dataset contains 2511 examples of hate/abusive speech.

\subsection{convAI}

\citet{curry} collect \emph{convAI} dataset from the user conversation with an AI assistant, CarbonBot, hosted on Facebook Messenger and a rule-based conversational agent, ELIZA. The authors categorize the dataset based on the severity and the type of abusiveness, directness, and target. We collected 853 examples from this dataset of which 128 are abusive speech.

\subsection{hateCheck}

\emph{hateCheck} is a synthetically-generated dataset collected from~\citet{rottger}. The authors develop 29 functionality through prior research and human interview and generate test case to evaluate test case for each of the functionalities. The dataset contains 2563 examples of hate speech.

\subsection{gab}

We collect \emph{gab} dataset from~\citet{qian}. Unlike other datasets,~\citet{qian} provide the full conversation which helps the models to understand the context. We collect 15,926 examples from the original corpus of which 15,270 are hate speech.

\subsection{yt-reddit}

\emph{yt-reddit} dataset is collected from~\citet{mollas}. The authors develop the dataset, namely, ETHOS sampling from Youtube and Reddit comments. The authors emphasize reducing any kinds of bias (e.g. gender) in the annotation process and annotate the dataset into various forms of targeted hate speech (e.g. origin, race, disability). We sample an equal number of hate and normal speech from this dataset.

\begin{figure}[t]
  \includegraphics[width=\linewidth]{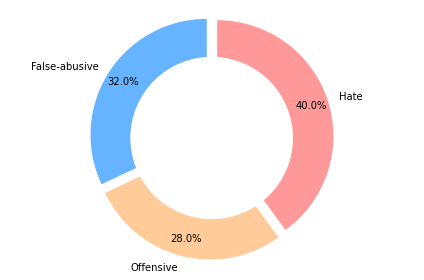}
  \caption{Percentage of error categories in a randomly sampled 50 misclassified examples.}
  \label{fig:error_analysis}
\end{figure}

\section{Error Analysis}
\label{appendix:error_analysis}

We conduct an error analysis on the examples that SCL-Fish misclassified. We randomly sample 50 misclassified examples and divide them into three categories:\\
\textbf{False-abusive:} Examples that are normal but SCL-Fish categorizes them as abusive.\\
\textbf{Offensive:} Examples that are degrading, harassing to an individual or untargeted abuse, trolling but SCL-Fish categorizes them as normal.\\
\textbf{Hate:} Examples that contain targeted attacks towards a particular group or identity.

\begin{table*}[t]

\centering

\resizebox{\linewidth}{!}{
\begin{tabular}{clc}
\hline
\textbf{Category}                    & \textbf{Example}                                                                                                                                        & \textbf{Predicted Class} \\ \hline
{\color[HTML]{3166FF} False-abusive} & \begin{tabular}[c]{@{}l@{}}Once at work, I accidentally checked out some dragons\\ f*cking cars. Didn't know that would be an issue.\end{tabular}       & Abusive                  \\
{\color[HTML]{FFCB2F} Offensive}     & The girl on the left looks like a Jewish pizza.                                                                                                         & Normal                   \\
{\color[HTML]{FE0000} Hate}          & \begin{tabular}[c]{@{}l@{}}The Blacks and Browns are destroying Sweden and nobody\\ does anything about it!\end{tabular}                                & Normal                   \\
{\color[HTML]{3166FF} False-abusive} & \begin{tabular}[c]{@{}l@{}}I've been out for years and it hasn't been easy, but I also\\ love being a f*g.\end{tabular}                                 & Abusive                  \\
{\color[HTML]{FFCB2F} Offensive}     & \begin{tabular}[c]{@{}l@{}}Why do you not think so? Feck off. Please go on. You annoy\\ me.\end{tabular}                                                & Normal                   \\
{\color[HTML]{FE0000} Hate}          & \begin{tabular}[c]{@{}l@{}}All we have is Blacks and a few Mestizos and all are pretty\\ well behaved, we keep them that way around here.\end{tabular}  & Normal                   \\
{\color[HTML]{3166FF} False-abusive} & I truly am a bada*s motherf*cker.                                                                                                                       & Abusive                  \\
{\color[HTML]{FFCB2F} Offensive}     & \begin{tabular}[c]{@{}l@{}}Oh yeah, she's so annoying I had mentally blocked out her\\ existence.\end{tabular}                                          & Normal                   \\
{\color[HTML]{FE0000} Hate}          & \begin{tabular}[c]{@{}l@{}}As opposed to Reddit, which is composed mainly of \\ Communists jacking each other off about child drag queens?\end{tabular} & Normal                   \\ \hline
\end{tabular}
}
\caption{\label{table:error_examples}
Sample of misclassified examples from \textit{False-abusive}, \textit{Offensive}, and \textit{Hate} category.
}
\end{table*}

 We provide examples for each category in Table~\ref{table:error_examples}. Figure~\ref{fig:error_analysis} shows that SCL-Fish misclassify 32\% normal examples as false-abusive. Most of the examples of this category contain some sort of slang words that the model confuses as abusive. On other hand, SCL-Fish misclassifies 28\% of the offensive examples as normal. This is because the examples may contain some positive words (e.g. `please') or do not contain any profanity. Therefore, the model considers them as normal speech. Lastly, around 40\% of the hate speech is misclassified as normal by SCL-Fish. Similar to the reason for offensive, the model confuse because of some sarcastic positive words and lack of expected profanity. This analysis shows that detecting implicit abusive language that does not contain direct profanity is still challenging and a direction to be explored in the future.

\end{document}